\documentclass[pdflatex,sn-mathphys-num]{sn-jnl}

\usepackage{graphicx}%
\usepackage{multirow}%
\usepackage{amsmath,amssymb,amsfonts}%
\usepackage{amsthm}%
\usepackage{mathrsfs}%
\usepackage[title]{appendix}%
\usepackage{xcolor}%
\usepackage{textcomp}%
\usepackage{manyfoot}%
\usepackage{booktabs}%
\usepackage{algorithm}%
\usepackage{algorithmicx}%
\usepackage{algpseudocode}%
\usepackage{listings}%
\usepackage{lmodern}
\usepackage{anyfontsize} 
\usepackage{placeins}
\begin{document}

\title[Article Title]{AMBER - Advanced SegFormer for Multi-Band Image Segmentation: an application to Hyperspectral Imaging}

\author*[1]{\fnm{Andrea} \sur{Dosi}}\email{andrea.dosi@unina.it}

\author[1,2,3]{\fnm{Massimo} \sur{Brescia}}\email{massimo.brescia@unina.it}
\equalcont{These authors contributed equally to this work.}

\author[2,3]{\fnm{Stefano} \sur{Cavuoti}}\email{stefano.cavuoti@inaf.it}
\equalcont{These authors contributed equally to this work.}

\author[4,2]{\fnm{Mariarca} \sur{D'Aniello}}\email{mariarca.daniello@unina.it}
\equalcont{These authors contributed equally to this work.}

\author[3]{\fnm{Michele} \sur{Delli Veneri}}\email{delliven@infn.it}
\equalcont{These authors contributed equally to this work.}

\author[4,5]{\fnm{Carlo} \sur{Donadio}}\email{donadio@unina.it}
\equalcont{These authors contributed equally to this work.}

\author[1]{\fnm{Adriano} \sur{Ettari}}\email{a.ettari@studenti.unina.it}
\equalcont{These authors contributed equally to this work.}

\author[1]{\fnm{Giuseppe} \sur{Longo}}\email{giuseppe.longo@unina.it}
\equalcont{These authors contributed equally to this work.}

\author[1]{\fnm{Alvi} \sur{Rownok}}\email{a.rownok@studenti.unina.it}
\equalcont{These authors contributed equally to this work.}

\author[1]{\fnm{Luca} \sur{Sannino}}\email{luca.sannino2@studenti.unina.it}
\equalcont{These authors contributed equally to this work.}

\author[1]{\fnm{Maria} \sur{Zampella}}\email{maria.zampella2@studenti.unina.it}
\equalcont{These authors contributed equally to this work.}

\affil*[1]{\orgdiv{Department of Physics E. Pancini}, \orgname{University of Naples Federico II}, \orgaddress{\street{Via Cinthia 21}, \city{Naples}, \postcode{80126} \country{Italy}}}

\affil[2]{\orgdiv{INAF - Osservatorio Astronomico di Capodimonte}, \orgaddress{\street{via Moiariello 16}, \city{Naples}, \postcode{80131}, \country{Italy}}}

\affil[3]{\orgdiv{INFN -- Section of Naples}, \orgaddress{\street{via Cinthia 9}, \city{Naples}, \postcode{80126}, \country{Italy}}}

\affil[4]{\orgdiv{Department of Earth Sciences, Environment and Resources}, \orgname{University of Naples Federico II}, \orgaddress{\street{Via Cinthia 21}, \city{Naples}, \postcode{80126}, \country{Italy}}}

\affil[5]{\orgdiv{Stazione Zoologica Anton Dohrn},  \orgaddress{\street{Villa Comunale}, \city{Naples}, \postcode{80121}, \country{Italy}}}

\abstract{Deep learning has revolutionized the field of hyperspectral image (HSI) analysis, enabling the extraction of complex spectral and spatial features. While convolutional neural networks (CNNs) have been the backbone of HSI classification, their limitations in capturing global contextual features have led to the exploration of Vision Transformers (ViTs). This paper introduces AMBER, an advanced SegFormer specifically designed for multi-band image segmentation. AMBER enhances the original SegFormer by incorporating three-dimensional convolutions, custom kernel sizes and a Funnelizer layer. This architecture enables to process hyperspectral data directly, without requiring spectral dimensionality reduction during preprocessing.
Our experiments, conducted on three benchmark datasets (Salinas, Indian Pines and Pavia University) and on a dataset from the PRISMA\footnote[1]{Project carried out using PRISMA Products, © of the Italian Space Agency (ASI), delivered under an ASI License to use.} satellite, show that AMBER outperforms traditional CNN-based methods in terms of Overall Accuracy, Kappa coefficient, and Average Accuracy on the first three datasets, and achieves state-of-the-art performance on the PRISMA dataset.
These findings highlight AMBER’s robustness, adaptability to both airborne and spaceborne data, and its potential as a powerful solution for remote sensing and other domains requiring advanced analysis of high-dimensional data.}


\maketitle

\section{Introduction}\label{sec1}
Hyperspectral imaging (HSI) captures detailed spectral information across a wide range of wavelengths, offering a unique and rich data source for various remote sensing applications, including environmental monitoring, agriculture, and mineral exploration. The classification of hyperspectral images involves identifying the material composition of each pixel, a task that requires complex algorithms capable of handling the high dimensionality and complexity of the data.
\newline
In recent years, deep learning algorithms, particularly convolutional neural networks (CNNs), have become prominent in HSI classification due to their ability to learn hierarchical features from raw data. CNN-based models have shown considerable success in extracting spectral-spatial features \cite{Survey1, ComparativeReview}. However, the inherent limitation of CNNs in capturing spectral-spatial long-range dependencies has spurred interest in transformer-based models {\cite{MIMO}.
\newline
Vision Transformers (ViTs) \cite{imageWord}, with their self-attention mechanism, have demonstrated remarkable success in various image-processing tasks. They are adept at capturing global contextual information, which is crucial for HSI classification. Many deep learning models have adopted the transformer architecture for hyperspectral data, and have achieved good results (e.g., \cite{spectralformer, LightWeightTransformer, Ahmad_2024, BUTT2024103773, Ghosh_2022, CasFormer, MSTNet}).
\newline
This paper introduces AMBER, an advanced version of SegFormer \cite{segformer}, specifically designed for multi-band image segmentation. AMBER incorporates several key innovations: it integrates three-dimensional convolutions, employs custom kernel sizes and strides to preserve output spatial dimensions. It also includes a Funnelizer layer to reduce spectral dimensions for generating two-dimensional masks. These enhancements enable AMBER to effectively process both spectral and spatial dimensions without sacrificing critical information.
\newline
We evaluated AMBER on four datasets—Salinas, Indian Pines, Pavia University, and PRISMA—where it demonstrated superior performance compared to traditional methods on the Salinas, Indian Pines and Pavia University datasets, achieving state-of-the-art results on the PRISMA dataset. Notably, AMBER exhibits robustness across diverse scenarios and conditions, performing well with data from both airborne spectrometers (AVIRIS for Salinas and Indian Pines; ROSIS for Pavia University) and spaceborne spectrometers (PRISMA). \newline
A distinctive feature of AMBER is its ability to handle hyperspectral data without requiring a preprocessing step to reduce spectral dimensions (e.g., SVD, or Gabor Filter \cite{gaborFilter, Dosi}), adopting a full-spectrum approach instead. This not only streamlines the workflow but also preserves essential spectral information, which is often crucial for accurate classification and analysis. The preservation of such spectral details is particularly beneficial for applications like land cover classification, mineral identification, and environmental monitoring, where subtle differences in spectral signatures can reveal significant insights. \newline
AMBER's innovative architecture significantly enhances representation learning by effectively addressing the unique challenges of hyperspectral data and, more broadly, complex multi-dimensional data cubes. Its ability to process such data makes it highly versatile, with potential applications extending beyond hyperspectral imaging. Future work will aim to broaden its applicability to additional domains, including medical imaging and astronomical data, where the analysis of multi-dimensional datasets plays a crucial role.

\section{Related Works}\label{sec2}
Deep learning algorithms have transformed the landscape of hyperspectral image (HSI) pixel classification by learning complex and meaningful features hierarchically. Among these, convolutional neural networks (CNNs) have become a cornerstone for exploring spectral-spatial features within local windows, owing to their robustness and scalability (\cite{Liu}).
\newline
In the last years, new CNN architectures have emerged with the goal of classifying/segmenting hyperspectral images. For example, Zhong, Li, Luo, and Chapman (2017) \cite{Zhong} introduced an end-to-end spectral-spatial residual network that employs residual blocks to continuously learn rich hyperspectral features. Paoletti, Haut, Fernandez-Beltran, et al. (2018) \cite{paolettiPiramidal} proposed a pyramidal residual network that incrementally diversifies high-level spectral-spatial attributes across layers. Zhu, Jiao, Liu, Yang, and Wang (2020) \cite{ZHAO} brought forth the residual spectral-spatial attention network (RSSAN), which cascades spectral and spatial attention modules for superior classification accuracy. Additionally, Chhapariya, Buddhiraju, and Kumar (2022) \cite{Chhapariya} designed a multi-level 3-D CNN with varied kernel sizes to extract multi-level features. Despite these advancements, CNNs still struggle to capture global contextual features and long-range dependencies for both hyperspectral and three-channel images due to their intrinsic locality.
\newline
\newline
The emergence of Vision Transformers (ViTs) is transforming image-processing tasks with their powerful modeling capabilities. Transformers have revolutionized natural language processing (NLP) and are now being increasingly used in various other fields, such as computer vision and hyperspectral imaging.
The key innovation of the transformer model is the self-attention mechanism \cite{Attention}, which allows the model to weigh the importance of different input tokens (words or image patches) when generating an output. This enables the transformer to capture long-range dependencies and relationships within the data. Vision Transformers adapt the transformer architecture, originally designed for sequential data like text, to process image data. In ViTs, an image is divided into fixed-size patches, which are then linearly embedded and treated as input tokens for the transformer model. In this context, SegFormer \cite{segformer} is an advanced vision transformer specifically designed for image segmentation tasks. It combines the strengths of transformers and CNNs, to effectively capture both global context and local details required for precise segmentation. 
\newline
This approach was successfully adapted to hyperspectral imaging in many transformer-based models. In particular SpectralFormer (Hong et al., 2021) \cite{spectralformer}, achieving impressive classification performance. Building on this foundation, Zhang, Meng, Zhao, Liu, and Chang (2022) \cite{ZHAO} emphasized the need for models that harness both global and local features. Sun, Zhao, Zheng, and Wu (2022) \cite{Sun} developed the spectral-spatial feature tokenization transformer (SSFTT) to capture high-level semantic and spectral-spatial features, demonstrating excellent performance across multiple HSI datasets. Mei, Song, Ma, and Xu (2022) \cite{Mei} proposed the group-aware hierarchical transformer (GAHT), which confines multi-head self-attention to local spatial-spectral contexts, thereby boosting classification accuracy. He et al. (2021) \cite{HeTransf} suggested a spectral-spatial transformer combining a sophisticated CNN with a dense transformer, effectively capturing spatial features and spectral relationships, leading to robust classification results.
\section{Methodology}\label{sec3}
This section introduces AMBER, a new and advanced SegFormer designed specifically for Multi-Band Image Segmentation. As the original SegFormer \cite{segformer}, AMBER consists of two main modules: a hierarchical Transformer encoder, which generates high-resolution coarse features and low-resolution fine features; and a lightweight All-MLP decoder, used to fuse multi-level features together. \newline
Unlike the original SegFormer, AMBER is tailored to handle multi-band images through the integration of three-dimensional convolutions. Furthermore, AMBER preserves the input spatial dimensions
$H$ and $W$ by employing custom kernel sizes and strides, ensuring higher accuracy. This enhancement comes with only a slight increase in the number of trainable parameters. A key addition in AMBER is the Funnelizer layer, which reduces the spectral dimensionality, enabling the generation of a two-dimensional segmentation mask from a three-dimensional input image. For an overview of the architecture, see Fig. \ref{amber_arch}.
\FloatBarrier
\begin{figure}[!ht]
\centering
\includegraphics[width=\textwidth]{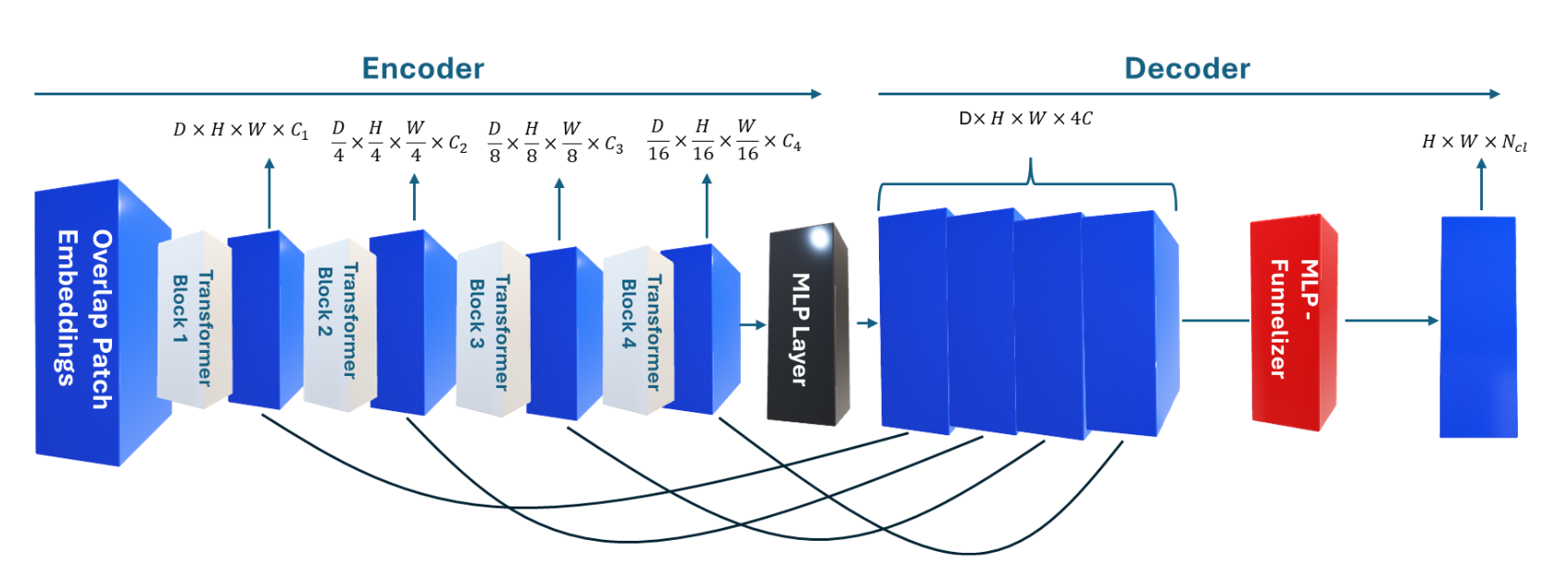}
\caption{The proposed AMBER framework consists of two main modules: A hierarchical Transformer encoder to extract spatial/spectral features; and a All-MLP decoder to fuse these multi-level features; the Funnelizer layer reduce the spectral dimension and predict the segmentation mask.}\label{amber_arch}
\end{figure}
\FloatBarrier

\noindent Given an image of $D\times H \times W$ (where $D$ represents the spectral dimension, $H$ denotes the height, and $W$ represents the width of the image), we first divide  it into patches of 
$3  \times 3 \times 3$ using a three-dimensional convolution layer (the patch size is the kernel dimension of the convolutional layer). We then
use these patches as input to the hierarchical Transformer encoder to obtain multi-level features at {1/4, 1/8, 1/16} of the original image resolution. We then pass these multi-level features to the All-MLP decoder to predict the segmentation mask at a $\frac{D}{4} \times \frac{H}{4} \times \frac{W}{4} \times N_{cls}$ resolution, where $N_{cls}$ is the number of classes. To provide a detailed explanation of our model architecture, we will follow almost the same steps as in the original paper \cite{segformer}.
\subsection{Hierarchical Transformer Encoder}
We create a series of Mix Transformer encoders (MiT) specifically designed for semantic segmentation of multi-band images, such as hyperspectral images.
\subsubsection*{Hierarchical Feature Representation.} Given an input image with a resolution of $D \times H \times W$, we perform patch merging to obtain a hierarchical feature map $F_{i}$ with a resolution of ${\frac{D}{2^{i+1}} \times \frac{H}{2^{i+1}} \times \frac{W}{2^{i+1}} \times C_{i}}$, where $i \in {1,2,3,4}$, and $C_{i+1}$ is larger than $C_{i}$.
\subsubsection*{Overlapped Patch Merging.} 
We utilize merging overlapping patches to avoid the need for positional encoding. To this end, we define $K$, $S$, and $P$, where $K$ is the three dimensional kernel size (or patch size), S is the stride between two adjacent patches, and $P$ is the padding size. Unlike the original SegFormer, in our experiments, we set $K = 3$, $S = 1$, $P = 1$ ,and $K = 3$, $S = 2$, $P = 2$ to perform overlapping patch merging. The patch size is intentionally kept small to preserve image details and avoid parameter explosion. $S = 1$ preserves the original image spatial dimensions $H$ and $W$, avoiding the reduction of spatial dimensions by $1/4$.
\subsubsection*{Efficient Self-Attention.}
In the multi-head self-attention process, each of the heads $Q,K,V$ have the same dimensions $N \times C$, where $N =D \times H \times W$ is the length of the sequence, the self-attention is estimated as: 
\begin{equation}
\text{Attention}(Q,K,V ) = \text{Softmax}\left( \frac{QK^{T}} {\sqrt{d_{head}}} \right)V.
\end{equation} We utilize the same approach as in the original paper to simplify the self-attention mechanism from a complexity of $O(N^{2})$ to $O(\frac{N^{2}}{R})$. In our experiments, we set $R$ to $[64, 16, 4, 1]$ from stage-1 to stage-4.
\subsubsection*{Mix-FFN.}
We introduce Mix-FFN \cite{segformer} which considers the effect of zero padding using a $3 \times 3 \times 3$ Conv in the feed-forward network (FFN). Mix-FFN can be formulated as: 
\begin{equation}
x_{out} = \text{MLP}(\text{GELU}(\text{Conv}_{3\times3\times3}(\text{MLP}(x_{in})))) + x_{in}
\end{equation}
where $x_{in}$ is the feature from the self-attention module. Mix-FFN mixes a $3 \times 3 \times 3$ convolution and an MLP into each FFN
\subsection{Lightweight All-MLP Decoder}
The All-MLP decoder consists of five main steps. First, multi-level features from the MiT encoder go through a MLP layer to unify the channel dimension. Then, in a second step, features are up-sampled to the original three-dimensional image and concatenated together. Third, a MLP layer is adopted to fuse the concatenated features. \newline 
\newline
The Funnelizer layer, a convolution with a kernel size of $(D \times 1 \times 1)$ reduces the spectral dimension, transforming the data into a two-dimensional representation. Finally, a two-dimensional convolutional layer with a kernel size of 1 is used to predict the segmentation mask, which has a resolution of $H \times W \times N_{cls}$, where $N_{cls}$ is the number of classes.
\section{Experiments and Analysis}\label{sec4}
\subsection{Data and experimental environment description}
In order to evaluate the proposed architecture, we utilize hyperspectral imaging, which records a detailed spectrum of light for each pixel and provides an invaluable source of information regarding the physical nature of the different materials, leading to the potential of a more accurate classification. Experiments were conducted firstly, on three of the most widely used public datasets — Salinas, Indian Pines, and Pavia University \cite{hyperDatasets}— as well as on a hyperspectral image acquired from the PRISMA satellite. Here is an overview of the datasets.
\subsubsection*{Salinas, Indian Pines and Pavia University datasets}
The Salinas, Indian Pines, and Pavia University datasets consist of hyperspectral images with water absorption bands removed for better spectral analysis.

\noindent The Salinas dataset was acquired using the AVIRIS spectrometer over the Salinas Valley, California. It is characterized by high spatial resolution (3.7-meter pixels). After the removal of water absorption bands, the image contains 204 spectral bands and has dimensions of $512 \times 217$ pixels.

\noindent The Indian Pines dataset was also captured by the AVIRIS spectrometer, covering a mixed agricultural and forested area in Northwestern Indiana, USA. The image has dimensions of $145 \times 145$ pixels, a spectral range of $0.40-2.50, \mu$m, 200 spectral bands, and a spatial resolution of 20 meters.

\noindent The Pavia University dataset was acquired using the ROSIS spectrometer during a flight campaign over the University of Pavia, Italy. This dataset features an image with dimensions of $610 \times 340$ pixels, a spectral range of $0.43-0.86, \mu$m, 103 spectral bands, and a spatial resolution of 1.3 meters.

\noindent The Salinas and Indian Pines datasets each contain 16 ground-truth classes, while the Pavia University dataset has 9 ground-truth classes. The false-color images and ground-truth maps for these datasets are shown in Figs. \ref{salinas_img_gt}, \ref{indianPines}, and \ref{paviaU}, with the class labels and their respective sample counts provided in Tables \ref{labelSalinas}, \ref{labelIndian} and \ref{labelPavia}.

\FloatBarrier
\begin{figure}[!ht]
\centering
\includegraphics[width=0.7\textwidth]{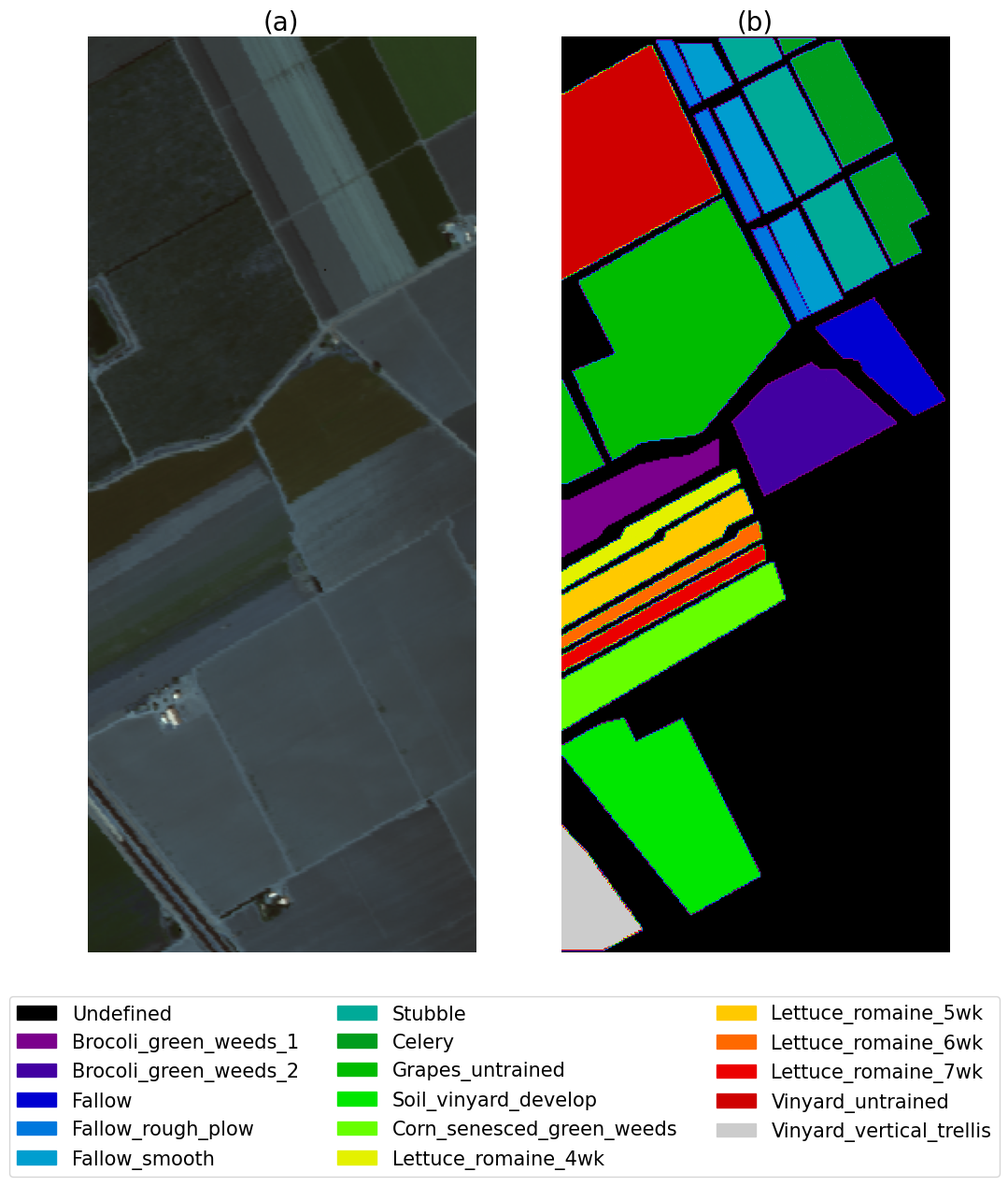}
\caption{Salinas dataset. (a) False-color map; (b) Ground-truth map.}\label{salinas_img_gt}
\end{figure}
\FloatBarrier

\FloatBarrier
\begin{table}[ht]
\caption{Groundtruth classes for the Salinas scene and their respective samples number.}\label{labelSalinas}%
\begin{tabular}{@{}llllll@{}}
\toprule
\textbf{\#} & \textbf{Class}& \textbf{Samples}\\
\midrule
1    & Brocoli green weeds 1   & 2009\\
2    & Brocoli green weeds 2 & 3726\\
3    & Fallow   & 1976\\
4    & Fallow rough plow  & 1394 \\
5    & Fallow smooth   & 2678 \\
6    & Stubble   & 3959 \\
7    & Celery   & 3579\\
8    & Grapes untrained   & 11271\\
9    & Soil vinyard develop   & 6203\\
10    & Corn senesced green weeds & 3278\\
11    & Lettuce romaine 4wk   & 1068\\
12    & Lettuce romaine 5wk & 1927\\
13    & Lettuce romaine 6wk   & 916\\
14    & Lettuce romaine 7wk   & 1070\\
15    & Vinyard untrained & 7268\\
16    & Vinyard vertical trellis & 1807\\
\botrule
\end{tabular}
\end{table}
\FloatBarrier

\FloatBarrier
\begin{figure}[!ht]
\centering
\includegraphics[width=\textwidth]{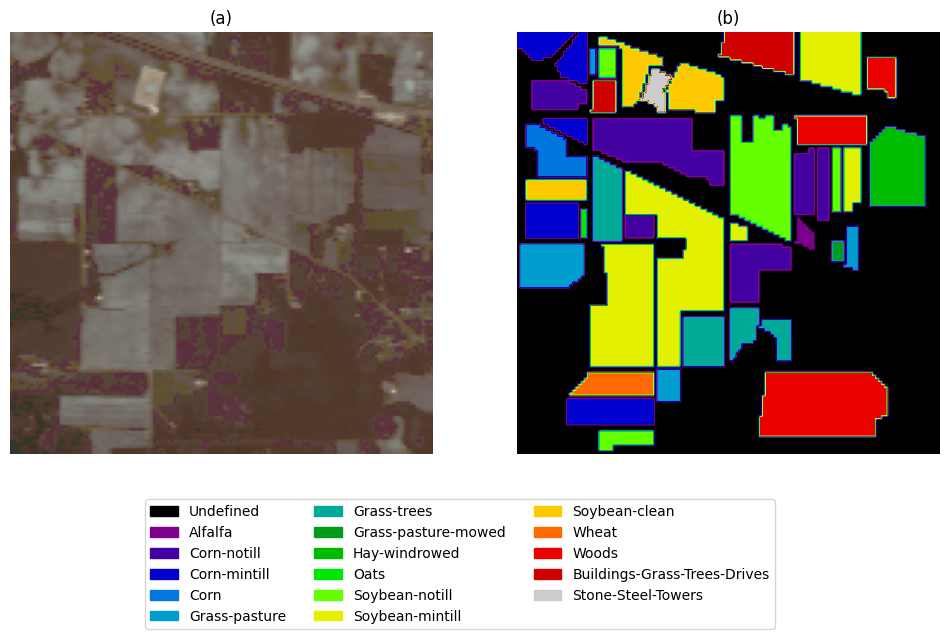}
\caption{Indian Pines dataset. (a) False-color map; (b) Ground-truth map.}\label{indianPines}
\end{figure}
\FloatBarrier

\FloatBarrier
\begin{table}[!ht]
\caption{Groundtruth classes for the Indian Pines scene and their respective samples number.}\label{labelIndian}%
\begin{tabular}{@{}llllll@{}}
\toprule
\textbf{\#} & \textbf{Class}& \textbf{Samples}\\
\midrule
1    & Alfalfa   & 46\\
2    & Corn-notill & 1428\\
3    & Corn-mintill   & 830\\
4    & Corn  & 237 \\
5    & Grass-pasture   & 483 \\
6    & Grass-trees   & 730 \\
7    & Grass-pasture-mowed   & 28\\
8    & Hay-windrowed   & 478\\
9    & Oats   & 20\\
10    & Soybean-notill & 972\\
11    & Soybean-mintill   & 2455\\
12    & Soybean-clean & 593\\
13    & Wheat   & 205\\
14    & Woods   & 1265\\
15    & Buildings-Grass-Trees-Drives & 386\\
16    & Stone-Steel-Towers & 93\\
\botrule
\end{tabular}
\end{table}
\FloatBarrier

\begin{figure}[!ht]
\centering
\includegraphics[scale=0.5]{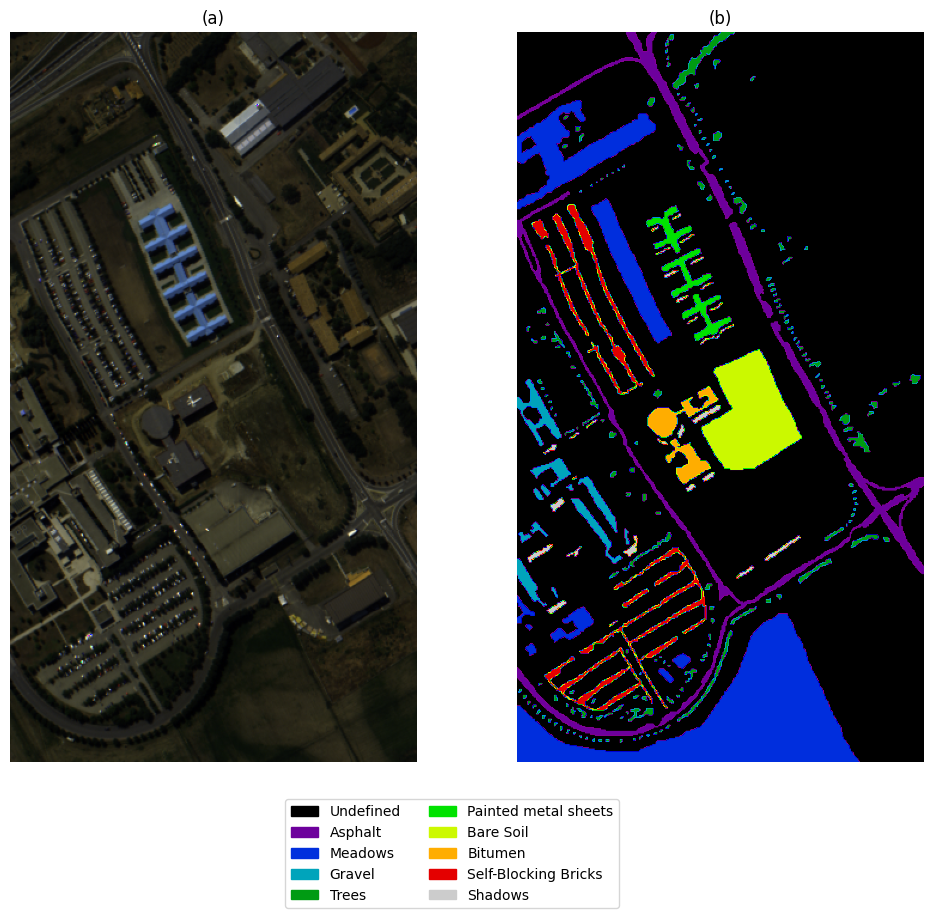}
\caption{Pavia University dataset. (a) False-color map; (b) Ground-truth map.}\label{paviaU}
\end{figure}
\FloatBarrier

\FloatBarrier
\begin{table}[ht]
\caption{Groundtruth classes for the Pavia University scene and their respective samples number.}\label{labelPavia}%
\begin{tabular}{@{}llllll@{}}
\toprule
\textbf{\#} & \textbf{Class}& \textbf{Samples}\\
\midrule
1    & Asphalt   & 6631\\
2    & Meadows & 18649\\
3    & Gravel   & 2099\\
4    & Trees  & 3064 \\
5    & Painted metal sheets   & 1345 \\
6    & Bare Soil   & 5029 \\
7    & Bitumen   & 1330\\
8    & Self-Blocking Bricks   & 3682\\
9    & Shadows   & 947\\
\botrule
\end{tabular}
\end{table}
\FloatBarrier

\subsubsection*{PRISMA dataset}
PRISMA, a space-borne hyperspectral sensor developed by the Italian Space Agency (ASI), can capture images in a continuum of 231 spectral bands ranging between $400$ and $2500$ nm, at a spatial resolution of 30 m. The method was applied to a PRISMA Bottom-Of-Atmosphere (BOA) reflectance scene of the Quadrilátero Ferrífero area (Minas Gerias State, Brazil), located in the central portion of the Minas Gerais State (Northern Brazil). \newline
The PRISMA hyperspectral image shows a reflectance scene at the Bottom-Of-Atmosphere (BOA) of the Quadrilátero Ferrífero mining district. It was acquired on May 21, 2022, at 13:05:52 UTC.\newline Level 2C PRISMA hyperspectral imagery was accessed from the mission website \url{http://prisma.asi.it/}. The VNIR and SWIR datacubes were pre-processed using tools available in ENVI 5.6.1 (L3Harris Technologies, USA). Errors in absolute geolocation were corrected using the Refine RPCs (HDF-EOS5) task, included within the PRISMA Toolkit in ENVI.
Only 193 out of 231 bands were used for the final image due to atmospheric absorption affecting the excluded bands.
\newline
The ESA WorldCover-Map, available at \url{https://esa-worldcover.org/en}, was used in conjunction with the Global-scale mining polygons (Version 1) \cite{globalMining} to establish the ground truth for the Brazilian region. In order to avoid redundant and noisy information, only the following features were selected: vegetation (including dense, sparse, and very sparse vegetation as well as crops), water areas, mining areas, and build-up areas. As it can be seen from Table \ref{labelPrisma}, the vegetation class in the image is over-represented. To address this issue, a random selection of $700,000$ vegetation pixels are set to 0 (undefined class). This adjustment ensures that the number of vegetation class pixels is comparable with the other classes. This is illustrated in Fig. \ref{prisma_image}.

\FloatBarrier
\begin{figure}[!ht]
\centering
\includegraphics[width=\textwidth]{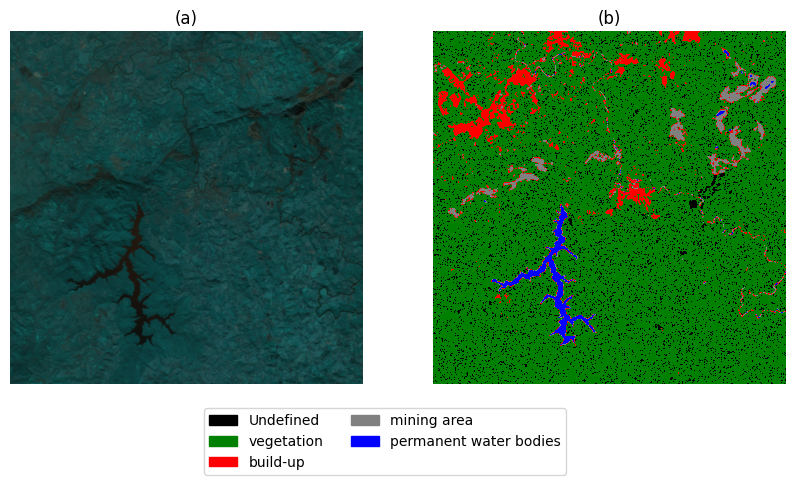}
\caption{PRISMA dataset. (a) False-color map; (b) Ground-truth map.}\label{prisma_image}
\end{figure}
\FloatBarrier
\begin{table}[ht]
\caption{Groundtruth classes for the PRISMA scene and their respective samples number.}\label{labelPrisma}%
\begin{tabular}{@{}llllll@{}}
\toprule
\textbf{\#} & \textbf{Class}& \textbf{Samples}\\
\midrule
1    & Vegetation   & 793572\\
2    & Mining Area & 36227\\
3    & Permanent Water bodies   & 19638\\
4    & Build-up  & 15463 \\
\botrule
\end{tabular}
\end{table}
\FloatBarrier

\subsection{Implementation details}
We trained AMBER and conducted inferences using the IBiSCo Data Center (Infrastructure for Big Data and Scientific Computing). Detailed information about the Data Center can be found at \url{https://ibiscohpc-wiki.scope.unina.it/}. Specifically, AMBER was trained using four Tesla V100 GPUs, each providing 32 GB of memory. Training times varied based on dataset size and complexity: approximately 6 hours for the Salinas dataset, 5 hours for the Indian Pines dataset, and around 7 hours for both the Pavia University and PRISMA datasets. Unlike the original SegFormer \cite{segformer}, AMBER was trained entirely from scratch, with no pre-training for the encoder and random initialization of the decoder.

\noindent During the training phase, we applied random flipping to the patches and excluded the undefined class from the computation of the loss function. The crop size was set to
$D \times 32 \times 32$ (where $D$ is the spectral dimension of the datasets) for all four datasets. This crop size was chosen after empirical evaluation: larger patches led to poorer predictions as AMBER struggled to capture finer details, while smaller patches resulted in a loss of context, degrading performance as well.

\noindent We trained the model using an SGD optimizer for all experiments,  with a batch size of 6 for Salinas, 3 for the Indian Pines dataset, 10 for the Pavia University dataset, and 4 for the PRISMA dataset. The batch size was selected based on the memory constraints of the GPUs and the dataset characteristics, ensuring stable and efficient training. The learning rate was set to a constant value of 0.01 across all experiments, as introducing a dynamic learning rate provided no significant benefits; the loss function consistently reached a plateau within the chosen number of epochs.

\noindent The model was trained for 30 epochs on the Salinas and Pavia University datasets, 50 epochs on the Indian Pines and PRISMA datasets, as these values ensured the loss function stabilized while avoiding overfitting. In all experiments, we used the Categorical Cross Entropy loss function to optimize performance.

\noindent To evaluate segmentation performance, we report metrics including Overall Accuracy (OA), Kappa Coefficient (Kappa), and Average Accuracy (AA). These metrics will be explained in detail in the following section.

\subsection{Evaluation metrics}
In order to effectively assess the performance of this method in the experiment, three indicators, Overall Accuracy (OA), Kappa Coefficient (Kappa) and Avarage Accuracy (AA), are used to evaluate the classification performance of the model. 
\newline
OA is calculated as the sum of correctly classified pixels divided by the total number of pixels. The number of correctly classified pixels is found along the diagonal of the confusion matrix, and the total number of pixels is equal to the total number of pixels of all real reference sources. The formula for OA is as follows:
\begin{equation}
 \text{OA} = \frac{\sum_{i=1}^{n} h_{ii}}{N} \times 100 / \% 
\end{equation}
Where:
\begin{itemize}
    \item $( h_{ii} )$ is the number of correctly classified pixels distributed along the diagonal of the confusion matrix.
    \item $N$ is the total number of samples.
    \item $n$ is the number of categories.
\end{itemize}
The Kappa coefficient is a measure of consistency in testing. It is calculated by multiplying the total number $(N)$ of all real reference pixels by the sum of the diagonal $( h_{kk} )$ of the confusion matrix. Then, the number of real reference pixels in each category and the classified pixels $( h_{ik}, h_{jk} )$ are subtracted from the product of the total number. This result is divided by the square of the total number of pixels minus the product of the total number of true reference pixels in each category and the total number of classified pixels in the category. The kappa coefficient comprehensively considers the various factors in the confusion matrix and provides a more comprehensive reflection of the accuracy of the overall classification. A larger value of the Kappa coefficient indicates higher accuracy of the corresponding classification algorithm. The general formula is as follows:
\begin{equation}
\text{Kappa} = \frac{N \sum_{k} h_{kk} - \sum_{k} \left( \sum_{j} h_{kj} \sum_{i} h_{ik} \right)}{N^2 - \sum_{k} \left( \sum_{j} h_{kj} \sum_{i} h_{ik} \right)}
\end{equation}
The Average Accuracy (AA) measures the average percentage of correctly classified samples for an individual class. It is calculated using the formula: 
\begin{equation}
    \text{AA} = \frac{\sum_{i=1}^{C} \left( {M_{ii}}/{\sum_{j=1}^{C} M_{ij}} \right)}{C} \times 100 / \%
\end{equation}
Where $C$ is the total number of labels or classes, and $M_{ij}$ represents the samples that actually belonged to the ith class and were predicted to belong to the $j-th$ class.

\subsection{Data Preprocessing}
We constructed the datasets using only random patches, in contrast to other studies that employ various data balancing strategies (e.g., \cite{PAOLETTI2018120}). This approach aimed to evaluate AMBER's ability to learn from unbalanced datasets and to create datasets with a consistent number of patches. The center of each patch was randomly selected from pixels that were not located on the edges of the images. For the Salinas, Indian Pines, and Pavia University datasets, the center was chosen from pixels that did not belong to the undefined class, ensuring patches contained meaningful information.

\noindent To mitigate the well-known issue of pixel overlap between training and test sets, which can lead to information leakage \cite{informationLeakage, Grewal2022HyperspectralIS}, we allocated a small percentage of each dataset for training. This methodology aligns with established practices in the literature (e.g, \cite{spectralformer, LiYing, Liang}). Specifically, for the Salinas, Indian Pines and Pavia University datasets, 20\% of the patches were assigned to the training set, while the remaining 80\% were used for testing. For the PRISMA dataset, the training set comprised only 10\% of the patches, with 90\% reserved for testing.

\noindent Before the training phase, dimensionality reduction techniques, such as PCA, SVD, or Gabor filters \cite{LiYing, gaborFilter, Dosi}, are commonly applied to the spectral dimension of hyperspectral images to address information redundancy. However, in our approach, we retained the full spectral dimension of the patches. This decision aimed to enable fine-grained classification and delegate the task of extracting relevant features to AMBER's encoder.

\section{Results}\label{sec5}
\subsection{Classification Results}
To assess the effectiveness of the proposed method in classifying hyperspectral remote sensing images, we compared it against four mainstream approaches: SVD/Unet \cite{Dosi}, HSI-CNN \cite{luo2018hsicnn}, 3D-CNN \cite{He}, and contextual deep CNN \cite{lee}. Notably, not all methods employ pixel-by-pixel classification. Among these methods, only SVD/Unet, contextual deep CNN, and AMBER utilize this approach, whereas HSI-CNN and 3D-CNN focus on image-level classification, where class labels are assigned based on the central pixel of each patch. This distinction inherently boosts the reported metrics for the latter two methods. Despite this, AMBER achieves superior performance on the Salinas, Indian Pines and Pavia University datasets and delivers results comparable to the best methods on the PRISMA dataset.

\noindent In the SVD/Unet method, the spectral dimension of the image is first reduced by a Singular Value Decomposition, considering only the first three singular values, and then a Unet architecture is used to classify pixels. This method has shown promising results in the semantic segmentation of PRISMA hyperspectral images.

\noindent In the HSI-CNN, the spectral-spatial features are first extracted from a target pixel and its neighbors. Next, several one-dimensional feature maps are obtained by performing a convolution operation on the spectral-spatial features, and these maps are then stacked into a two-dimensional matrix. Finally, this two-dimensional matrix, treated as an image, is input into a standard CNN.

\noindent The3D-CNNarchitecture is a Multiscale 3D deep convolutional neural network (M3DDCNN) which could jointly learn both 2D Multi-scale spatial feature and 1D spectral feature from HSI data in an end-to-end approach.

\noindent The contextual deep CNN first concurrently applies multiple 3-dimensional local convolutional filters with different sizes jointly exploiting spatial and spectral features of a hyperspectral image. The initial spatial and spectral feature maps obtained from applying the variable size convolutional filters are then combined together to form a joint spatio-spectral feature map. The joint feature map representing rich spectral and spatial properties of the hyperspectral image is then fed through fully convolutional layers that eventually predict the corresponding label of each pixel vector.

\noindent Each neural network mentioned above was trained on the same patches using the network settings and configurations described in the corresponding papers \cite{Dosi, luo2018hsicnn, He, lee}. In this context, ``same patches" refers to patches centered on the same pixel. AMBER was trained on patches of dimension $D \times 32 \times 32$, where $D$ represents the spectral dimension, as previously mentioned. SVD/Unet was trained on patches of $3 \times 32 \times 32$ (3 corresponding to the top three singular values), HSI-CNN on patches of $D \times 3 \times 3$, 3D-CNN on patches of $D \times 7 \times 7$, and contextual deep CNN on patches of $D \times 5 \times 5$. As described in most articles (e.g., \cite{LiYing, Liang, Chen, spectralformer}), the datasets were divided into training and testing sets by random sampling throughout the image, ensuring that the samples did not overlap. In this context, ``samples" refer to the central pixel of the patch.

\noindent For all methods, we conducted five independent experiments on each dataset. In each experiment, we used five-fold Monte Carlo cross-validation, using various random sampling of the entire image to generate different training and test sets \cite{Nalepa_2019}. This approach ensured robustness and reliability in evaluating the model's performance.

\noindent The results presented in Tables \ref{tab1}, \ref{tab2}, \ref{tab3}, and \ref{tab4} represent the mean classification accuracy for each class, along with the mean Overall Accuracy (OA), Kappa coefficient, and Average Accuracy (AA) computed across the five-fold Monte Carlo cross-validation sets, together with their standard deviations (±) to illustrate the variability and reliability of the model's performance across different validation folds. The highest values for each metric are highlighted in bold.

\FloatBarrier
\begin{table}[ht]
\caption{Classification accuracy of Salinas Dataset.}\label{tab1}%
\begin{tabular}{@{}llllll@{}}
\toprule
Class & SVD/Unet \cite{Dosi} & HSI-CNN \cite{luo2018hsicnn}  & 3D-CNN \cite{He} & Cont Deep CNN \cite{lee} & \textbf{AMBER}\\
\midrule
1    & $0.00$   & $99.10$  & $99.49$ & $97.14$ & $\boldsymbol{99.95}$ \\
2    & $99.99$   & $97.29$  & $99.99$ & $99.82$ & $\boldsymbol{100}$ \\
3    & $34.24$   & $84.52$  & $\boldsymbol{99.97}$ & $99.31$ & $99.65$ \\
4    & $91.47$   & $96.91$  & $98.60$ & $99.52$ & $\boldsymbol{99.68}$ \\
5    & $99.55$   & $98.49$  & $98.09$ & $94.17$ & $\boldsymbol{99.65}$ \\
6    & $99.96$   & $99.98$  & $\boldsymbol{100}$ & $99.99$ & $99.99$ \\
7    & $65.26$   & $99.69$  & $99.96$ & $\boldsymbol{99.99}$ & $99.98$ \\
8    & $97.96$   & $86.49$  & $96.09$ & $90.95$ & $\boldsymbol{99.46}$ \\
9    & $84.33$   & $99.12$  & $99.85$ & $\boldsymbol{99.95}$ & $96.99$ \\
10    & $79.15$   & $91.84$  & $96.76$ & $95.63$ & $\boldsymbol{99.74}$ \\
11    & $31.48$   & $96.92$  & $98.91$ & $95.98$ & $\boldsymbol{99.63}$ \\
12    & $32.15$   & $\boldsymbol{100}$  & $99.37$ & $99.92$ & ${99.93}$ \\
13    & $99.57$   & $98.90$  & $95.29$ & $\boldsymbol{99.67}$ & $99.57$ \\
14    & $99.05$   & $95.56$  & $\boldsymbol{99.83}$ & $95.52$ & $99.16$ \\
15    & $0.013$   & $56.94$  & $77.89$ & $78.32$ & $\boldsymbol{97.60}$ \\
16    & $64.54$   & $96.34$  & $99.11$ & $99.16$ & $\boldsymbol{99.97}$ \\
OA/\%    & $72.28 \pm 2.94$   & $90.20 \pm 0.65$  & $96.08 \pm 0.36$ & $84.42 \pm 2.02$ & $\boldsymbol{99.73 \pm 0.053}$ \\
Kappa × 100    & $67.87 \pm 3.42$   & $88.99 \pm 0.029$  & $95.59 \pm 0.42$ & $81.05 \pm 2.56$ & $\boldsymbol{99.69 \pm 0.061}$  \\
AA/\%   & $67.42 \pm 6.80$   & $93.63 \pm 0.24$  & $97.46 \pm 0.19$ & $71.40 \pm 9.02$ & $\boldsymbol{99.75 \pm 0.021}$  \\
\botrule
\end{tabular}
\end{table}
\FloatBarrier

\FloatBarrier
\begin{table}[ht]
\caption{Classification accuracy of Indian Pines Dataset.}\label{tab2}%
\begin{tabular}{@{}llllll@{}}
\toprule
Class & SVD/Unet \cite{Dosi} & HSI-CNN \cite{luo2018hsicnn}  & 3D-CNN \cite{He} & Cont Deep CNN \cite{lee} & \textbf{AMBER}\\
\midrule
1    & $24.56$   & $18.28$  & $94.01$ & $83.79$ & $\boldsymbol{99.27}$ \\
2    & $\boldsymbol{99.70}$   & $30.78$  & $93.58$ & $77.10$ & $99.52$ \\
3    & $99.36$   & $5.29$  & $83.44$ & $22.10$ & $\boldsymbol{99.42}$ \\
4    & $57.24$   & $7.66$  & $77.85$ & $48.69$ & $\boldsymbol{99.49}$ \\
5    & $96.27$   & $51.96$  & $94.94$ & $71.89$ & $\boldsymbol{99.16}$ \\
6    & $99.96$   & $96.09$  & $99.69$ & $99.25$ & $\boldsymbol{99.98}$ \\
7    & $0.00$   & $0.00$  & $94.25$ & $86.79$ & $\boldsymbol{96.02}$ \\
8    & $91.98$   & $95.30$  & $99.63$ & $93.49$ & $\boldsymbol{99.97}$ \\
9    & $0.0029$   & $1.56$  & $93.65$ & $34.85$ & $\boldsymbol{96.95}$ \\
10    & $75.17$   & $52.75$  & $88.94$ & $87.74$ & $\boldsymbol{99.35}$ \\
11    & $99.93$   & $89.80$  & $95.83$ & $89.77$ & $\boldsymbol{99.86}$ \\
12    & $94.022$   & $21.35$  & $88.38$ & $49.17$ & $\boldsymbol{99.16}$ \\
13    & $73.19$   & $97.36$  & $\boldsymbol{100}$ & $98.46$ & $99.95$ \\
14    & $99.97$   & $95.49$  & $99.83$ & $99.19$ & $\boldsymbol{99.97}$ \\
15    & $72.0055$   & $32.03$  & $96.22$ & $60.46$ & $\boldsymbol{97.42}$ \\
16    & $73.58$   & $79.07$  & $98.94$ & $99.90$ & $\boldsymbol{99.54}$ \\
OA/\%    & $94.18 \pm 4.22$   & $68.15 \pm 7.36$  & $95.01 \pm 0.79$ & $84.42 \pm 2.02$ & $\boldsymbol{99.74 \pm 0.097}$ \\
Kappa × 100    & $92.77 \pm 5.40$   & $60.55 \pm 10.029$  & $94.03 \pm 0.94$ & $81.05 \pm 2.56$ & $\boldsymbol{99.68 \pm 0.12}$  \\
AA/\%   & $70.99 \pm 4.22$   & $47.059 \pm 9.28$  & $93.25 \pm 1.10$ & $71.40 \pm 9.02$ & $\boldsymbol{99.18 \pm 0.34}$  \\
\botrule
\end{tabular}
\end{table}
\FloatBarrier

\FloatBarrier
\begin{table}[ht]
\caption{Classification accuracy of Pavia University Dataset.}\label{tab3}%
\begin{tabular}{@{}llllll@{}}
\toprule
Class & SVD/Unet \cite{Dosi} & HSI-CNN \cite{luo2018hsicnn}  & 3D-CNN \cite{He} & Cont Deep CNN \cite{lee} & \textbf{AMBER}\\
\midrule
1    & $89.19$   & $96.79$  & $98.94$ & $97.48$ & $\boldsymbol{99.79}$ \\
2    & $96.47$   & $99.09$  & $99.37$ & $99.51$ & $\boldsymbol{100}$ \\
3    & $0.17$   & $74.44$  & $90.70$ & $82.06$ & $\boldsymbol{99.27}$ \\
4    & $95.18$   & $94.58$  & $98.87$ & $97.21$ & $\boldsymbol{99.60}$ \\
5    & $94.15$   & $99.89$  & $\boldsymbol{100}$ & $99.98$ & $99.99$ \\
6    & $91.50$   & $98.14$  & $97.81$ & $95.05$ & $\boldsymbol{99.99}$ \\
7    & $44.09$   & $96.90$  & $97.06$ & $97.12$ & $\boldsymbol{99.88}$ \\
8    & $\boldsymbol{99.91}$   & $96.34$  & $98.29$ & $96.05$ & $99.85$ \\
9    & $99.75$   & $99.70$  & $99.86$ & $99.96$ & $\boldsymbol{99.98}$ \\
OA/\%    & $91.12 \pm 8.2$   & $96.96 \pm 0.77$  & $98.51 \pm 0.31$ & $97.40 \pm 0.84$ & $\boldsymbol{99.94 \pm 0.0098}$ \\
Kappa × 100    & $86.72 \pm 11.82$   & $96.15 \pm 0.98$  & $98.12 \pm 0.39$ & $96.49 \pm 1.14$ & $\boldsymbol{99.91 \pm 0.015}$  \\
AA/\%   & $70.99 \pm 4.22$   & $95.11 \pm 1.66$  & $97.87 \pm 0.40$ & $96.04 \pm 0.64$ & $\boldsymbol{99.82 \pm 0.022}$  \\
\botrule
\end{tabular}
\end{table}
\FloatBarrier
\FloatBarrier
\FloatBarrier
\begin{table}[ht]
\caption{Classification accuracy of PRISMA Dataset.}\label{tab4}%
\begin{tabular}{@{}llllll@{}}
\toprule
Class & SVD/Unet \cite{Dosi} & HSI-CNN \cite{luo2018hsicnn}  & 3D-CNN \cite{He} & Cont Deep CNN \cite{lee} & \textbf{AMBER}\\

\midrule
1    & $90.57$   & $\boldsymbol{95.34}$  & $93.59$ & $77.97$ & $81.70$ \\
2    & $94.99$   & $68.97$  & $81.75$ & $89.52$ & $\boldsymbol{96.60}$ \\
3    & $\boldsymbol{92.65}$   & $74.32$  & $78.69$ & $87.80$ & $92.21$ \\
4    & $\boldsymbol{96.71}$   & $80.35$  & $83.71$ & $90.54$ & $93.18$ \\
OA/\%    & $\boldsymbol{93.52 \pm 0.090}$   & $89.88 \pm 3.57$  & $88.11 \pm 0.21$ & $88.52  \pm 0.26$ & $90.90 \pm 0.84$ \\
Kappa × 100    & $\boldsymbol{91.06 \pm 0.12}$   & $74.62 \pm 0.99$  & $80.46 \pm 0.51$ & $84.06 \pm 0.34$ & $87.41 \pm 1.48$  \\
AA/\%   & $\boldsymbol{93.73 \pm 0.10}$   & $79.74 \pm 0.69$  & $84.44 \pm 0.68$ & $87.66 \pm 0.54$ & $90.92 \pm 1.32$  \\
\botrule
\end{tabular}
\end{table}
\FloatBarrier

\FloatBarrier
\begin{figure}[!ht]
\centering
\includegraphics[width=\textwidth]{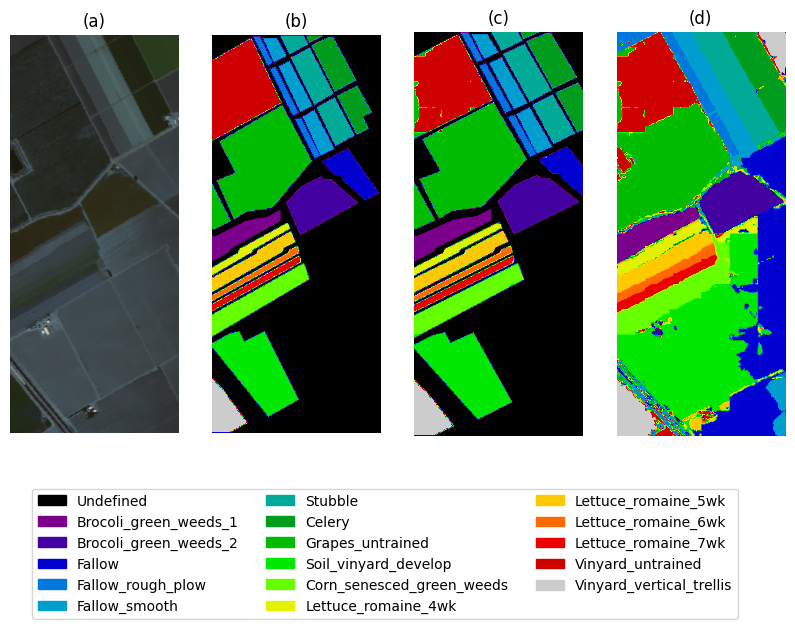}
\caption{Salinas prediction. (a) False-color map. (b) Ground-truth map. (c) AMBER prediction with the undefined mask. (d) AMBER prediction without the undefined mask.}\label{salias_pred}
\end{figure}
\FloatBarrier

\FloatBarrier
\begin{figure}[!ht]
\centering
\includegraphics[width=\textwidth]{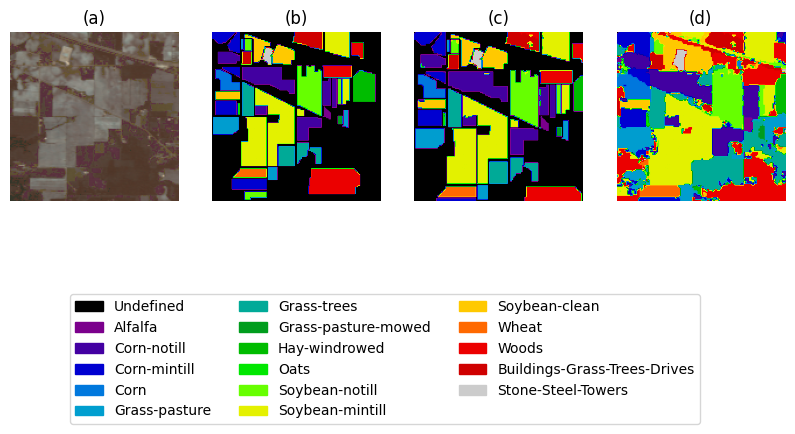}
\caption{Indian Pines prediction. (a) False-color map. (b) Ground-truth map. (c) AMBER prediction with the undefined mask. (d) AMBER prediction without the undefined mask.}\label{indian_image_pred}
\end{figure}
\FloatBarrier

\FloatBarrier
\begin{figure}[!ht]
\centering
\includegraphics[width=\textwidth]{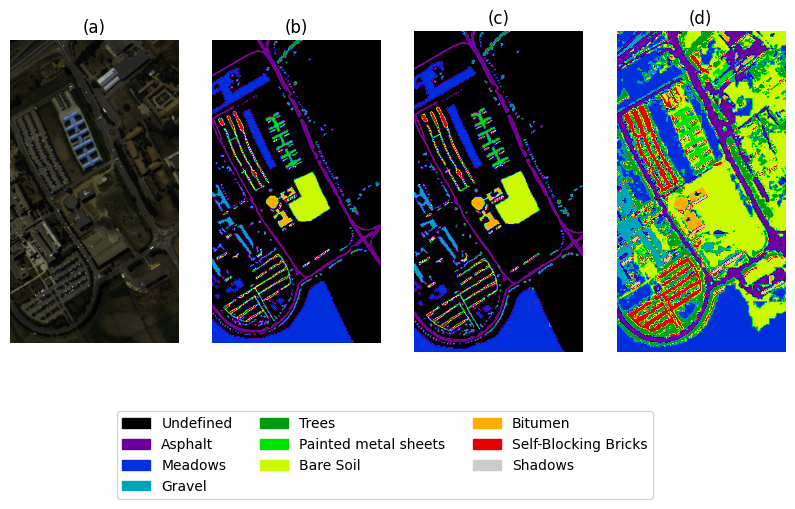}
\caption{Pavia University prediction. (a) False-color map. (b) Ground-truth map. (c) AMBER prediction with the undefined mask. (d) AMBER prediction without the undefined mask.}\label{pavia_image_pred}
\end{figure}
\FloatBarrier

\FloatBarrier
\begin{figure}[!ht]
\centering
\includegraphics[width=\textwidth]{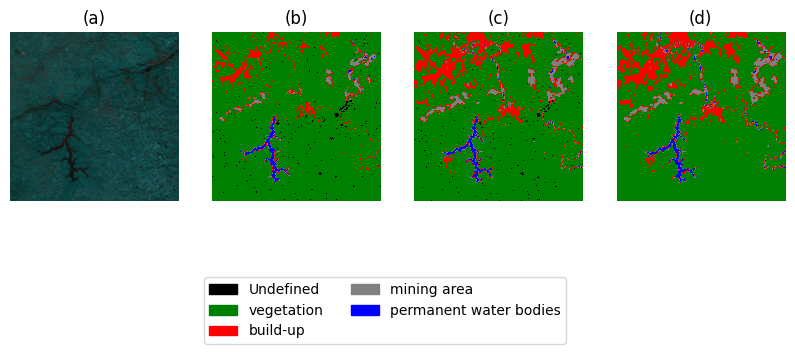}
\caption{PRISMA prediction. (a) False-color map. (b) Ground-truth map. (c) AMBER prediction with the undefined mask. (d) AMBER prediction without the undefined mask.}\label{prisma_image_pred}
\end{figure}
\FloatBarrier
\noindent The AMBER method consistently outperformed the compared methods on the Salinas, Indian Pines, and Pavia University datasets, achieving the highest Overall Accuracy (OA), Kappa coefficients (Kappa) and Average Accuracy (AA) across five independent experiments.

\noindent For the Salinas dataset (Table \ref{tab1}), AMBER achieved the best OA of $99.73\% \pm 0.053$, the best Kappa coefficient of $99.69 \pm 0.061$, and the best AA of $99.75\% \pm 0.021$. These metrics surpass all competing methods, including 3D-CNN ($96.08\% \pm 0.36$ OA) and HSI-CNN ($90.20\% \pm 0.65$ OA). AMBER outperformed other networks in 10 out of the 16 classes, achieving near-perfect accuracy in classes such as Class 2 ($100\%$) and Class 6 ($99.99\%$). Furthermore, AMBER demonstrated its robustness in minority classes like Class 15, achieving $97.60\%$, significantly higher than HSI-CNN ($56.94\%$) and SVD/Unet ($0.013\%$).

\noindent For the Indian Pines dataset (Table \ref{tab2}), AMBER again outperformed all other models, achieving an OA of $99.74\% \pm 0.097$, Kappa coefficient of $99.68 \pm 0.12$, and AA of $99.18\% \pm 0.34$. AMBER demonstrated exceptional accuracy across underrepresented classes, such as Class 7 (Grass-pasture-mowed), where it achieved $96.02\%$, while other methods like HSI-CNN and SVD/Unet failed entirely, with $0\%$ accuracy. This ability to handle class imbalance highlights AMBER's robustness. AMBER also achieved the highest or near-highest accuracy in most other classes, such as Class 13 (Soybean-mintill) and Class 14 (Soybean-clean), where it delivered $99.95\%$ and $99.97\%$, respectively.

\noindent For the Pavia University dataset (Table \ref{tab3}), AMBER achieved the best OA ($99.94\% \pm 0.0098$), Kappa coefficient ($99.91 \pm 0.015$), and AA ($99.82\% \pm 0.022$). It consistently delivered the highest accuracy across most classes, including minority classes like Class 3 (Gravel), where it achieved $99.27\%$, significantly outperforming SVD/Unet ($0.17\%$). AMBER also achieved near-perfect accuracy in dominant classes, such as Class 2 (Asphalt) and Class 9 (Shadows), achieving $100\%$ and $99.98\%$, respectively. Additionally, its low standard deviations across all metrics underscore its reliability and generalization capabilities.

\noindent For the PRISMA dataset (Table \ref{tab4}), AMBER achieved an OA of $90.90\%$, a Kappa coefficient of $0.8741$, and an AA of $90.92\%$. While these results were slightly lower than the SVD/Unet method, AMBER still outperformed other competing approaches. The marginally lower performance is under investigation, suggesting potential areas for further model optimization.

\noindent AMBER demonstrates exceptional performance across hyperspectral datasets, outperforming state-of-the-art methods such as 3D-CNN, HSI-CNN, and SVD/Unet on the Salinas, Indian Pines, and Pavia University datasets. It consistently achieves high accuracy in both majority and minority classes, effectively classifying challenging features such as Broccoli-green weeds (Salinas), Alfalfa, Grass-pasture-mowed, and Soybean-mintill (Indian Pines), and Asphalt, Gravel, and Bitumen (Pavia University). On the PRISMA dataset, AMBER achieves state-of-the-art performance, successfully identifying complex features such as mining areas and built-up regions. Its ability to generalize across both airborne datasets (Salinas, Indian Pines, Pavia University) and spaceborne data (PRISMA) is evident from its consistently high accuracy and low standard deviations in OA, Kappa and AA, underscoring its robustness and reliability.

\noindent The Fig. \ref{salias_pred}, \ref{indian_image_pred}, \ref{pavia_image_pred}, and \ref{prisma_image_pred} provide visual results for each dataset, including the false-color image (a), ground-truth sample plots (b), and AMBER's classification outputs (c) and (d). In particular, plot (c) shows the predicted classification results, while plot (d) overlays a mask of ground-truth undefined pixels to visually justify AMBER's high accuracy across all feature types presented in Tables \ref{tab1}, \ref{tab2}, \ref{tab3}, and \ref{tab4}. Notably, for the Salinas, Pavia University, and Indian Pines datasets, the ground-truth map (b) and AMBER's predictions (c) are nearly identical.

\noindent Figure \ref{pavia_patch} highlights AMBER's ability to accurately predict pixels classified as undefined in the ground-truth map, which are excluded from the loss function computation. The figure illustrates AMBER's capacity to segment complex structures. The example shows an asphalt class structure in the Pavia University dataset.
\FloatBarrier
\begin{figure}[!ht]
\centering
\includegraphics[width=\textwidth]{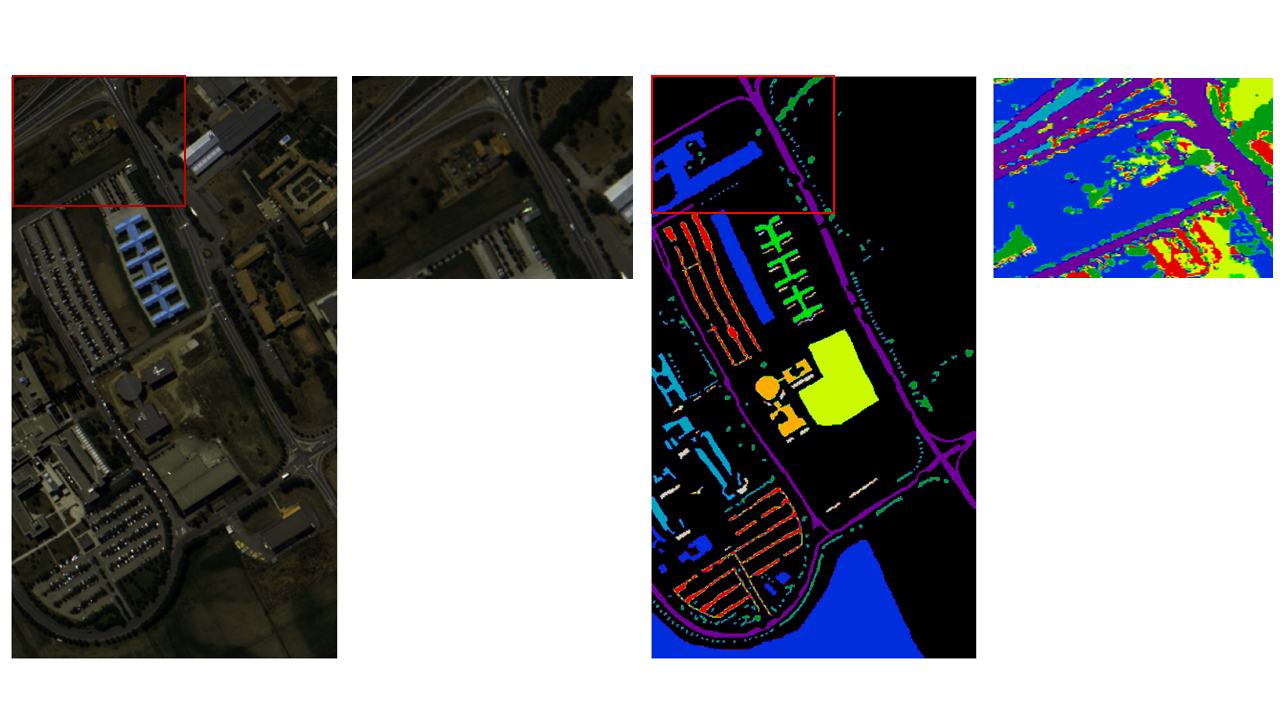}
\caption{AMBER prediction for the undefined pixels: a complex asphalt structure classification example}\label{pavia_patch}
\end{figure}
\FloatBarrier

\section{Conclusion}\label{sec6}
In this paper, we presented AMBER, an advanced SegFormer designed and fine-tuned for multi-band image segmentation. By incorporating three-dimensional convolutions and preserving spatial dimensions, AMBER effectively captures both local and global features in multi-band images. Extensive experiments on the Salinas, Indian Pines, Pavia University, and PRISMA datasets demonstrate that AMBER, in most cases, outperforms traditional CNN-based methods in terms of Overall Accuracy, Kappa coefficient, and Average Accuracy. \newline
The results indicate that AMBER significantly improves classification accuracy while exhibiting robust generalization capabilities across diverse hyperspectral datasets. Compared to traditional CNN methods like 3D-CNN and contextual deep CNN, AMBER's transformer-based architecture gives it a clear advantage by effectively integrating spectral and spatial features. This aligns with findings from prior studies that emphasize the importance of capturing global context in hyperspectral image analysis. Additionally, the segmentation performance on the PRISMA dataset highlights AMBER’s adaptability to satellite-derived hyperspectral data, which are often characterized by high spectral variability and noise. \newline
Despite these advancements, some challenges remain. For example, while AMBER achieved superior accuracy on most datasets, the performance on certain complex spectral regions suggests potential limitations in handling extreme spectral variations. This opens avenues for further research, particularly in enhancing the model's robustness through advanced regularization techniques or adaptive attention mechanisms. \newline
In a broader context, the adoption of transformer-based architectures like AMBER signals a paradigm shift in hyperspectral image analysis. The ability to process spatial and spectral information simultaneously makes these models well-suited for a variety of remote sensing applications, such as land-cover classification, environmental monitoring, and disaster assessment. Future research could extend the capabilities of AMBER by integrating domain-specific knowledge, leveraging transfer learning, and incorporating more diverse datasets. \newline
Moreover, AMBER addresses the challenge of segmenting astronomical objects, such as galaxies, by effectively processing multi-band images like radio interferometric data. Its ability to handle complex, high-dimensional datasets makes it a promising tool for advancing segmentation tasks in astronomy. Simultaneously, AMBER can be employed to segment three-dimensional medical images, such as Magnetic Resonance Imaging (MRI) or Computed Tomography (CT) scans, demonstrating its versatility in tackling diverse segmentation challenges across multiple domains. \newline
In summary, AMBER offers a robust and scalable approach to hyperspectral image segmentation, setting the stage for future innovations in the field. By building upon this work, researchers can further refine transformer-based methods to tackle emerging challenges and expand their applicability across diverse scientific domains.

\maketitle
\subsubsection*{Acknowledgements}
This work has been funded by project code PIR01\_00011 “IBISCo”, PON 2014-2020, for all three entities (INFN, UNINA and CNR) and by the PNRR CN1 - ISP\_S2\_AGR\@INTESA for UNINA.

\bibliography{sn-bibliography_new}

\end{document}